\documentclass[journal]{IEEEtran}

\usepackage{bigstrut,array,multirow,tabularx}
\usepackage{diagbox}
\usepackage{slashbox}
\usepackage{makecell}
\usepackage{amsthm}
\usepackage{tikz}
\usepackage[super]{nth}
\usepackage{algorithm}
\usepackage{algorithmic}
\usepackage{graphicx}
\usepackage{verbatim}
\usepackage{subfigure}  
\usepackage{adjustbox}
\def\BigRoman{\uppercase\expandafter{\romannumeral\number\count 255 }}
\def\Romannumeral{\afterassignment\BigRoman\count255=}

\usepackage{amsmath,cite}

\usepackage{amssymb}
\usepackage{amsthm}
\usepackage[normalem]{ulem}
\usepackage[hidelinks]{hyperref}
\usepackage[nameinlink, capitalise, noabbrev]{cleveref}
\usepackage{mathtools}

\newcommand{\RNum}[1]{\uppercase\expandafter{\romannumeral #1\relax}}

\newtheorem{problem}{Problem}

\ifCLASSINFOpdf
\else
\fi

\begin{document}
\title{Two-Stage Deep Anomaly Detection with Heterogeneous Time Series Data}

\author{Kyeong-Joong~Jeong,
        Jin-Duk~Park,
        Kyusoon~Hwang,
        Seong-Lyun~Kim,\\
        and~Won-Yong~Shin,~\IEEEmembership{Senior Member,~IEEE}%
        \thanks{This work was supported by the Republic of Korea’s MSIT (Ministry of Science and ICT), under the High-Potential Individuals Global Training Program (No. 2020-0-01463) supervised by the IITP (Institute of Information and Communications Technology Planning Evaluation) and by the Yonsei University Research Fund of 2020 (2020-22-0101). {\em (Corresponding author: Won-Yong Shin.)}}%
\thanks{K.-J. Jeong, J.-D. Park, and W.-Y. Shin are with the Department of Computational Science and Engineering, Yonsei University, Seoul 03722, Republic of Korea (e-mail: \{jeongkj,~jindeok6,~wy.shin\}@yonsei.ac.kr).}% 
\thanks{K. Hwang is with Wizcore Co., Ltd., Seoul 04209, Republic of Korea (e-mail: hwkyso@wizcore.co.kr).}% <-this % stops a space
\thanks{S.-L. Kim is with the Department of Electrical and Electronic Engineering, Yonsei University, Seoul 03722, Republic of Korea (e-mail: slkim@yonsei.ac.kr).}}% <-this % stops a space

\maketitle


\begin{abstract}

We introduce a data-driven anomaly detection framework using a manufacturing dataset collected from a factory assembly line. Given \emph{heterogeneous} time series data consisting of operation cycle signals and sensor signals, we aim at discovering abnormal events. Motivated by our empirical findings that conventional single-stage benchmark approaches may not exhibit satisfactory performance under our challenging circumstances, \color{black}we propose a \emph{two-stage} deep anomaly detection (T-DAD) framework in which two different unsupervised learning models are adopted depending on types of signals. In Stage \RNum{1}, we select anomaly candidates by using a model trained by operation cycle signals; in Stage \RNum{2}, we finally detect abnormal events out of the candidates by using another model, which is suitable for taking advantage of temporal continuity, trained by sensor signals. A distinguishable feature of our framework is that operation cycle signals are exploited first to find likely anomalous points, whereas sensor signals are leveraged to filter out unlikely anomalous points afterward. \color{black}Our experiments comprehensively demonstrate the superiority over single-stage benchmark approaches, the model-agnostic property, and the robustness to difficult situations.\color{black}

\end{abstract}


%\titlepgskip=-15pt

\maketitle

% \section*{Acknowledgment}
% This research project was supported by The Sports Promotion Fund of Seoul Olympic Sports Promotion Foundation From Ministry of Culture, Sports and Tourism (No.1375027368) and by the National Research Foundation of Korea (NRF) grant funded by the Korea government (MSIT) (No. 2021R1A2C3004345).

\section{Introduction}
\label{sec:introduction}

\subsection{Background}
\IEEEPARstart{W}{ith} the rapid development of the fourth industrial revolution (Industry 4.0), \color{black}which encompasses Industrial Internet of Things (IIoT) as well as smart and sustainable manufacturing systems~\cite{khakbaz2021sustainable}\color{black}, a vast amount of data are generated \color{black}every day\color{black}~\cite{7857790}. Along with the growing importance of Big Data analysis, studies on anomaly detection~\cite{chandola2009anomaly} have attracted increasing attention for modern industry. Anomaly detection enables us to solve a variety of important problems in smart industrial systems, e.g., security protection~\cite{8106743}, fault detection~\cite{8710319, ince2016real, 8565906}, and condition monitoring~\cite{9016153}. 

On the other hand, artificial intelligence (AI) or machine learning (ML) models have been widely used in discovering anomalies on industrial datasets~(see, e.g.,~\cite{8106743,8710319,8370640,9016153,ince2016real, 8565906, wang2016self,nakazawa2019anomaly,yin2020anomaly,malhotra2016lstm,wu2019lstm,audibert2020usad} and references therein). Nevertheless, \color{black}real-world \color{black}industrial datasets contain the following challenging issues~\cite{saufi2019challenges}: 1) the quality of data depends heavily on data acquisition environments; 2) observed patterns of anomalies may not be readily predictable; 3) and very few (trusted) labels are available. \color{black}To tackle such challenges, we should design AI/ML-based anomaly detection models in a data-driven fashion, which are capable of capturing \color{black}complex temporal or organic relations and being robust to both label noise and class imbalance. 

\subsection{Motivation and Main Contributions}

In this paper, we present a novel \color{black}data-driven \color{black}unsupervised approach for anomaly detection using a manufacturing dataset collected from a real-world factory assembly line, which accompanies a variety of considerably challenging problems such as severe data imbalance with very scarce labels and intrinsically unpredictable patterns. In particular, given \emph{heterogeneous} multivariate time series data consisting of \color{black}operation cycle \color{black}signals and \color{black}sensor \color{black}signals, we are interested in automatically discovering true abnormal events in the factory assembly line, which are defined as anomalies in our context.

Our study is motivated by our empirical findings that conventional single-stage benchmark detection approaches (even sophisticated) may not provide satisfactory performance under our real-world manufacturing dataset; this is because existing approaches are not inherently designed to suit our own dataset composed of two different signals (i.e., operation cycle signals and sensor signals), exhibiting significantly different patterns from each other. More precisely, when there are two different signals recorded over a multi-scale time span as in our dataset, existing anomaly detection methods are not adequate for making use of such signals altogether. As a novel and effective solution to resolve this issue, \color{black}we propose a \emph{two-stage} deep anomaly detection (T-DAD) framework in which two different unsupervised learning models are adopted depending on types of signals. The two models are trained without labels since unsupervised learning is rather appropriate for our dataset in which very few abnormal events are only available, similarly as in many studies in such industrial systems~\cite{9016153, ren2019time}. More specifically, in Stage \RNum{1}, we select anomaly candidates by using a model trained by using operation cycle signals; in Stage \RNum{2}, we finally detect abnormal events (i.e., the most likely anomalies) among the candidates by using another model, which is suitable for capturing the temporal dependence, trained only by using sensor signals. \color{black}A distinguishable feature of our \color{black}T-DAD \color{black}framework is that \emph{signal-specific} detection tasks are carried out in such a way that, along with the two empirically optimized thresholds, operation cycle signals are exploited first to find \color{black}likely \color{black}anomalous points whereas sensor signals are leveraged as a secondary tool to filter out \color{black}unlikely \color{black}anomalous points \color{black}afterward. \color{black}In other words, our \color{black}T-DAD \color{black}framework fully takes advantage of the characteristics of our heterogeneous time series data in detecting anomalies, which basically differs from prior attempts in the literature.

Through intensive experiments, we demonstrate the superiority of our \color{black}T-DAD \color{black}framework over various single-stage benchmark methods using a certain assembly process dataset in terms of the \color{black}best $F_1$ score\color{black}, representing the best possible performance of a trained model on a test set given the optimal threshold(s), where the $F_1$ score is the harmonic mean of \color{black}Precision \color{black}and \color{black}Recall\color{black}. \color{black}We also show the model-agnostic property of our T-DAD framework by diversifying two appropriately chosen learning methods in Stages \RNum{1} and \RNum{2} and then successfully integrating them into our framework\color{black}. To validate the robustness of the proposed \color{black}T-DAD \color{black}framework to other fringe environments, we evaluate the performance of anomaly detection using other assembly process datasets, which display consistent trends in terms of not only a balanced number of features in each of operation cycle and sensor signals but also a non-negligible fraction of anomalies by showing substantial gains of \color{black}T-DAD \color{black}over benchmark methods. 

Our methodology sheds crucial insights into a simple implementation of data-driven anomaly detection methods when \emph{heterogeneous} multivariate signals with unpredictable patterns are given as input and/or \color{black}very few (or no) labels \color{black}are available. 
\color{black}
Various problems in real-world manufacturing domains include but are not limited to fault detection, condition monitoring, manufacturing inspection, functional failure detection, alarm management, and so forth. When such practical problems deal with heterogeneous signals, using our distinctive two-stage strategy can be promising with potential gains. Moreover, as operational benefits in manufacturing domains, automatic discovery of abnormal events with much higher accuracy through our approach would enable cost-effective management in the Industry 4.0 era along with minimal or complementary interventions by the operator with domain knowledge.

\color{black}

\subsection{Organization}
The remainder of this paper is organized as follows. In Section \ref{Sec:2}, we summarize studies related to our work. Section \ref{Sec:3} describes our dataset. The overall methodology of \color{black}T-DAD \color{black}is presented in Section \ref{Sec:4}. Experimental results are provided in Section \ref{Sec:5}. In Section \ref{Sec:6}, we summarize the paper with some concluding remarks.
\par Table~\ref{summary of notations} summarizes the notations used in this paper, which will be formally defined in the following sections when we introduce our problem definition and technical details.

% \begin{comment}
\begin{table}[t]
\renewcommand{\arraystretch}{1.0}
\caption{Summary of notations}
\label{summary of notations}
\centering
\begin{tabular}{|c | c|}
\hline
\textbf{Notation} & \textbf{Description} \\ \hline

 ${{\bf x}_t^{(o)}}$ & Operation cycle signal at timestamp $t$  \\ 
  \hline
 ${{\bf x}_t^{(s)}}$ & Sensor signal at timestamp $t$ \\ 
% \hline
% $\mathcal{T}_o$ & a set of timestamps at which there exist operation cycle signals.\\ 
 \hline
 $a_t^{(o)}$ & Anomaly score from ${{\bf x}_t^{(o)}}$ \\
 \hline
 $a_t^{(s)}$& Anomaly score from ${{\bf x}_t^{(s)}}$ \\
 \hline
  $\delta$ & Tolerance \\ 
 \hline
 $\mathcal{T}_o$ & Set of timestamps where operation cycle signals exist \\
 \hline
  $\hat{\mathcal{T}}_{\text{anomaly}}$ & Set of detected abnormal events within the tolerance $\delta$ \\
 \hline
  $\tau_1$ & Threshold on Stage I \\ 
 \hline
 $\tau_2$ & Threshold on Stage II \\
  \hline
\end{tabular}
\end{table}
% \end{comment}

\begin{figure*}[t]
    \centering
    \includegraphics[width=1.0\textwidth]{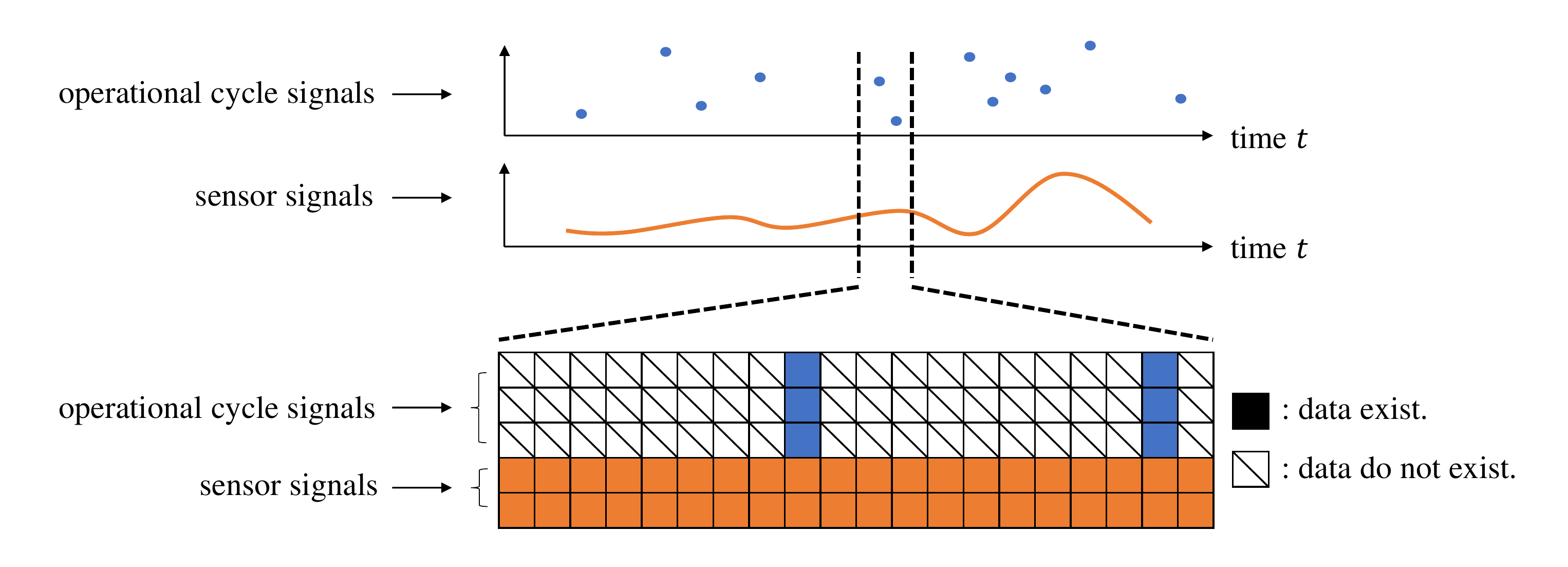}
    \caption{Heterogeneous features, including operation cycle signals and sensor signals.}
    \label{fig:heterofigure}
\end{figure*}

\section{Related Work}
\label{Sec:2}

The framework that we propose in this paper is related to two broader fields of research, namely 1) anomaly detection in real-world manufacturing domains 2) time series anomaly detection.

\subsection{Anomaly Detection in Manufacturing.}
    There has been a steady push to solve anomaly detection tasks using real-world manufacturing datasets such as real-time motor fault detection using motor current signals \cite{ince2016real}, \color{black}fault detection using wind turbine datasets \cite{8565906}, \color{black}condition monitoring using a simulated bearing fault dataset \cite{9016153}, manufacturing inspection using tile production data \cite{8370640}, and functional failure detection using sensor data collected from a turbine syngas compressor in global manufacturing plants \cite{8710319}. Anomaly detection in manufacturing or production lines (e.g., Intel's CPU manufacturing \cite{wang2016self}, and semiconductor manufacturing \cite{nakazawa2019anomaly}) has also received significant attention. 
    
%     %iiot sensors
\color{black}
\subsection{Time Series Anomaly Detection}
    
    Anomaly detection in a time series using stochastic time series models such as autoregressive moving average (ARMA) \cite{galeano2006outlier} and autoregressive integrated moving average (ARIMA) \cite{zhang2005network} has been actively studied in the past two decades. One-class support vector machine (SVM) was also widely used for time series novelty (i.e., out-of-distribution) detection \cite{ma2003time}. Moreover, probabilistic methods such as hidden Markov models \cite{fuse2017statistical}, Bayesian networks \cite{tartakovsky2012efficient}, and Poisson models \cite{9445591} have been presented for time series anomaly detection. 
    
    With recent advances in deep learning across a wide range of domains, anomaly detection tasks have also successfully employed various deep neural networks such as recurrent neural network (RNN), convolutional neural network (CNN), autoencoder (AE), variational autoencoder (VAE), generative adversarial network (GAN) and graph neural network (GNN). First, RNN models have been actively developed in time series anomaly detection to capture the temporal dependence in time series sequential data. The strength of using a long short term memory (LSTM) network for time series anomaly detection was demonstrated in \cite{lstmndt} by presenting a nonparametric and dynamic error thresholding approach that does not depend on either scare labels or prior distributional assumptions. A stochastic RNN for multivariate time series anomaly detection to learn robust representations with a stochastic variable connection and planar normalizing flow was presented in \cite{omnianomaly}. Second, with regard to CNN-based anomaly detection models, a deep CNN approach was developed in \cite{munir2018deepant} as a time series predictor module. Moreover, an algorithm that combines spectral residual and CNN was designed in \cite{ren2019time} to achieve the state-of-the-art performance on univariate time series anomaly detection. Third, AE-aided approaches have also been widely employed by taking into account the reconstruction error as an anomaly score. An AE-based algorithm was proposed in \cite{borghesi2019anomaly} for anomaly detection in high-performance computing (HPC) systems. A robust anomaly detection method integrating AE and robust principal component analysis was presented in \cite{zhou2017anomaly}. An encoder--decoder architecture-based anomaly detection method using LSTM layers was proposed in \cite{malhotra2016lstm}. As an end-to-end learning solution, an unsupervised anomaly detection model that utilizes the AE and Gaussian mixture model was proposed in \cite{dagmm}. In \cite{audibert2020usad}, a scalable anomaly detection method for multivariate time series was built upon an encoder--decoder architecture with an adversely training framework inspired by GAN. Fourth, among generative models, VAE or GAN-based time series anomaly detection approaches also received considerable attention. A VAE-based anomaly detection algorithm was designed in \cite{xu2018unsupervised} along with a solid theoretical explanation for adopting VAE on univariate time series anomaly detection. An approach combining LSTM and VAE was presented in \cite{lstmvae} for time series anomaly detection. An unsupervised anomaly detection method was also presented in \cite{madgan} based on GANs using LSTM networks for each generator and discriminator. Fifth, increasing attention has recently been paid to time series anomaly detection models \cite{zhao2020multivariate, gdn} based on GNNs \cite{wu2019survey} in order to capture complex relationships between variables in multivariate time series data.

\subsection{Discussions}
    As shown above, various time series anomaly detection approaches from traditional methods to current deep learning-based methods have been extensively developed in the literature. Nonetheless, to the best of our knowledge, there is no prior work on anomaly detection that makes use of the characteristics of heterogeneous time series signals similarly as in our real-world manufacturing dataset.

\color{black}

\begin{figure*}
    \centering
    \includegraphics[width=1.0\textwidth]{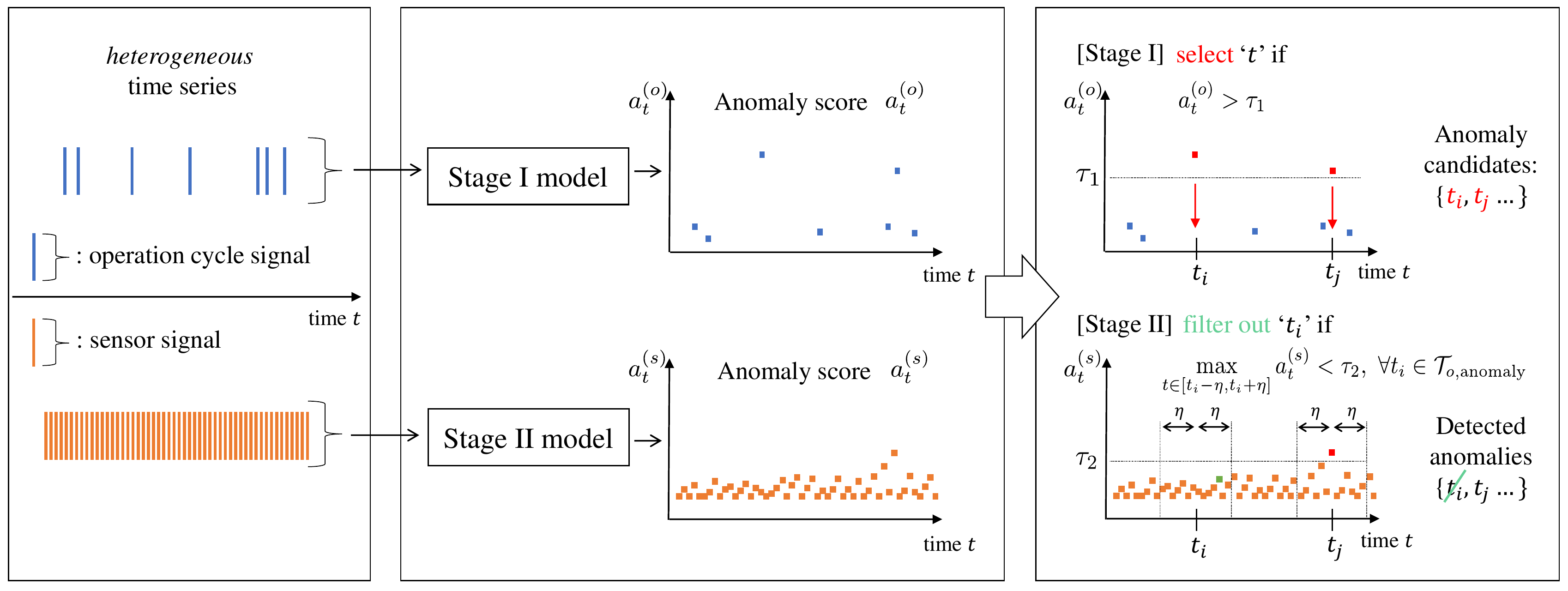}
    \caption{The schematic overview of our T-DAD framework.}
    \label{fig:overview}
\end{figure*}

\section{Dataset Description}
\label{Sec:3}

We use a multivariate time series dataset collected from an \color{black}assembly line of automotive engine valves \color{black}as one of the large-scale real-world manufacturing datasets. The entire dataset consists of a set of features recorded from December 2, 2019 to December 7, 2019, in which there are 518,400 timestamps (i.e., temporal records). \color{black}Let us describe the set of features, which are partitioned into the following two types of signals.\color{black}

\begin{itemize}
\item \emph{Sensor} signals: Since sensor signals are recorded every second regardless of the operation status, they tend to exhibit temporal continuity;
\item \emph{Operation cycle} signals: Due to the fact that operation cycle signals are logged in units of operation cycles, they are less likely to follow temporal continuity. Different from sensor signals, since such operation cycle signals are purely event-driven, they occur irregularly.
\end{itemize}
\color{black}
We note that sensor signals are evenly spaced in time while operation cycle signals are unevenly spaced in time; this comes from the fact that the operation is inactive when either the planned short downtime or the irregular (unplanned) long downtime occurs. The behavior of heterogeneous features is illustrated in Fig. \ref{fig:heterofigure}. 
\color{black}

\section{Methodology}
\label{Sec:4}
In this section, to describe the proposed methodology, we first present the problem definition. Then, we elaborate on our \color{black}T-DAD \color{black}framework.

\subsection{Problem Definition}

We define our problem in automatically detecting true abnormal events that have occurred in the factory assembly line using the heterogeneous time series manufacturing dataset. It would be senseless to perform \color{black}point-wise \color{black}anomaly detection since it is rather crucial for the operator to be aware of whether a possible abnormal situation has occurred within an acceptable time range~\cite{ren2019time}. Thus, we aim at conducting \color{black}range-wise \color{black}detection of abnormal events by discovering a \color{black}most likely anomalous point\color{black}, which is defined as a nearby timestamp manifesting an abnormal pattern from the point at which an abnormal event has indeed occurred.

\begin{problem}
Given a heterogeneous multivariate time series dataset, range-wise anomaly detection is to find an anomalous point $t$ such that there exists an abnormal event in $[t-\delta,t+\delta]$ for a certain tolerance $\delta>0$.
\end{problem}
\noindent Note that the tolerance $\delta$ is assumed to be set appropriately depending on the operator's criterion beforehand.

\subsection{Overall Architecture}

To accomplish anomaly detection with our heterogeneous multivariate time series data, we propose novel two-stage detection, termed T-DAD, built upon a data-driven approach inspired by non-straightforward empirical findings. \color{black}Fig. \ref{fig:overview} illustrates the overall architecture of the proposed T-DAD framework in which signal-specific detection tasks are carried out after training two different unsupervised learning models, the so-called Stage \RNum{1} and Stage \RNum{2} models, according to two types of signals (i.e., operation cycle signals and sensor signals). To be specific, we start by adopting the Stage \RNum{1} model with multivariate operation cycle signals and the Stage \RNum{2} model with multivariate sensor signals that account for the temporal dynamics, where each model is unsupervisedly trained without any labels. \color{black}Then, we compute anomaly scores, denoted by $a_t^{(o)}$ and $a_t^{(s)}$, from the output of the each model, respectively, at each timestamp $t$. Thereafter, we perform two-stage anomaly detection as follows. In Stage \RNum{1}, we select a set of \color{black}very likely anomaly candidates\color{black}, denoted by $\mathcal{T}_{o,\text{anomaly}}$, at which the resulting anomaly score $a_t^{(o)}$ is \color{black}higher \color{black}than a certain threshold $\tau_1 > 0$, i.e.,
\begin{align}
\mathcal{T}_{o,\text{anomaly}}=\left\{t | \forall t\in \mathcal{T}_o:~a_t^{(o)}>\tau_1\right\}, \label{EQ:1}
\end{align}
where $\mathcal{T}_o$ is the set of timestamps at which there exist operation cycle signals. On the contrary, in Stage \RNum{2}, we detect the most likely anomalous points by filtering out \color{black}very unlikely \color{black}anomalous points among the selected candidates. More precisely, we remove the set of points, $\mathcal{T}_s$, such that the maximum of the resulting anomaly scores $a_t^{(s)}$, $\forall t \in [t_i-\eta,t_i+\eta]$, is \color{black}lower \color{black}than another threshold $\tau_2>0$, where $t_i\in \mathcal{T}_{o,\text{anomaly}}$ and $\eta>0$ is a range setting parameter to be empirically found. That is, we finally detect the set of anomalies, $\hat{\mathcal{T}}_{\text{anomaly}}$, as in the following:
\begin{align}
\hat{\mathcal{T}}_{\text{anomaly}}=\mathcal{T}_{o,\text{anomaly}}\setminus \mathcal{T}_s, \label{EQ:2}
\end{align}
where
\begin{align}
\mathcal{T}_s=\Big\{t_i| \forall t_i \in \mathcal{T}_{o,\text{anomaly}}:~\max_{t\in [t_i-\eta,t_i+\eta]} a_t^{(s)}<\tau_2\Big\}. \label{EQ:3}
\end{align}

\subsection{Implementation of the T-DAD Framework}
In this subsection, we elaborate on the proposed \color{black}T-DAD \color{black}framework, which constitutes two models along with \color{black}signal-specific \color{black}detection tasks. In our study, we adopt the DAE model~\cite{borghesi2019anomaly,zhou2017anomaly} in Stage \RNum{1} with operation cycle signals and the LSTM-DAE model~\cite{malhotra2016lstm} in Stage \RNum{2} with sensor signals. \color{black}Note that, due to the model-agnostic property of our proposed framework, any unsupervised learning model with which the anomaly score is computed at each timestamp can be employed in each stage. Nevertheless, we focus on describing our T-DAD framework built upon the DAE and LSTM-DAE models since they are relatively tractable.\color{black}\footnote{\color{black}In Section V-F3, we shall carry out comprehensive experimental evaluations by diversifying unsupervised learning models in Stages \RNum{1} and \RNum{2}, which demonstrates that the use of DAE and LSTM-DAE in Stages \RNum{1} and \RNum{2}, respectively, indeed leads to the highest best $F_1$ score.\color{black}}

Let us describe the model training process. We first adopt a DAE model employing a feed-forward multi-layer neural network in which the desired output is the input itself, as stated in~\cite{borghesi2019anomaly,zhou2017anomaly}. Our DAE model contains a multi-layer perceptron (MLP) encoder and an MLP decoder, denoted by $E_{\text{MLP}}(\cdot)$ and $D_{\text{MLP}}(\cdot)$, respectively, where the multivariate operation cycle signal ${\bf x}_t^{(o)} \in \mathbb{R}^{1\times d_o}$ is used as input. Here, $d_o$ denotes the number of features in the operation cycle signal. By training this DAE, we are capable of measuring the reconstruction error at each timestamp $t \in \mathcal{T}_o$, which is regarded as the anomaly score $a_t^{(o)}=\|{\bf x}_t^{(o)}-\hat{\bf x}_t^{(o)}\|_2$, where $\hat{\bf x}_t^{(o)} = D_{\text{MLP}}(E_{\text{MLP}}({\bf x}_t^{(o)}))$ and $\|\cdot\|_2$ denotes the $\ell_2$ norm. 

In the meantime, to take advantage of temporal continuity, we adopt an LSTM-DAE model~\cite{malhotra2016lstm} along with the multivariate sensor signal ${\bf x}_t^{(s)} \in \mathbb{R}^{1\times d_s}$, where $d_s$ denotes the number of features in the sensor signal. This LSTM-DAE contains an encoder--decoder with LSTM layers, denoted by $E_{\text{LSTM}}(\cdot)$ and $D_{\text{LSTM}}(\cdot)$, respectively. We apply a sliding window technique with the window size $w>0$ and the step size $\gamma>0$ to the sensor signal ${\bf x}_t^{(s)}$ so as to generate sequences $S_1,S_2,\cdots$, which are used as input of LSTM-DAE, where $\gamma<w$ is assumed and $S_j=\{ {\bf x}_{1+(j-1)\gamma}^{(s)},\cdots, {\bf x}_{w+(j-1)\gamma}^{(s)} \}$ for $j \in\{1,2,\cdots\}$. Since a sensor signal ${\bf x}_t^{(s)}$ at timestamp $t$ is contained in multiple sequences, the reconstruction error at timestamp $t$ from a sequence $S_j$ can be expressed as $\| {\bf x}_t^{(s)} - \hat{\bf x}_{t;j}^{(s)} \|_2$, where $\hat{\bf x}_{t;j}^{(s)}$ is the entry of $D_{\text{LSTM}}(E_{\text{LSTM}}(S_j))$ corresponding to ${\bf x}_t^{(s)}$. We then take the maximum of those multiple reconstruction errors calculated at each timestamp $t$,\footnote{Other values such as the minimum and arithmetic mean can also be adopted. However, we have empirically verified that taking the maximum leads to the highest performance.} which is regarded as the anomaly score $a_t^{(s)} = \max_{j \in J_t} \| {\bf x}_t^{(s)} - \hat{\bf x}_{t;j}^{(s)} \|_2$, where $J_t$ is the set of sequences to which a sensor signal ${\bf x}_t^{(s)}$ at timestamp $t$ belongs. In consequence, heterogeneous multivariate time series signals \{${\bf x}_t^{(o)}, {\bf x}_t^{(s)}\}$ can be seen as univariate anomaly scores $\{a_t^{(o)}, a_t^{(s)}\}$ via two trained DAE models. 

\begin{figure}[t]
  \centering
  \subfigure[Fragement 1~(05:30:22--05:50:22 on December 3, 2019)] {\includegraphics[width=0.84\linewidth]{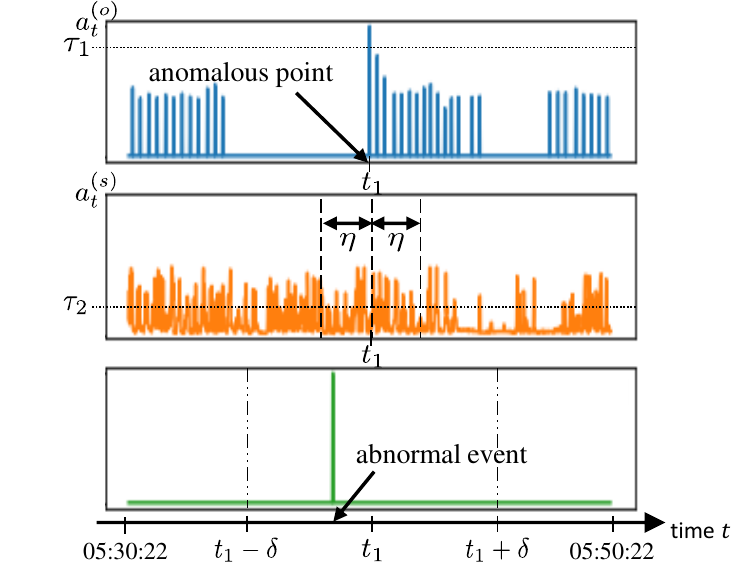}\label{fig:fig_2a}}
  \subfigure[Fragement 2~(04:36:15--04:56:15 on December 7, 2019)]{\includegraphics[width=0.84\linewidth]{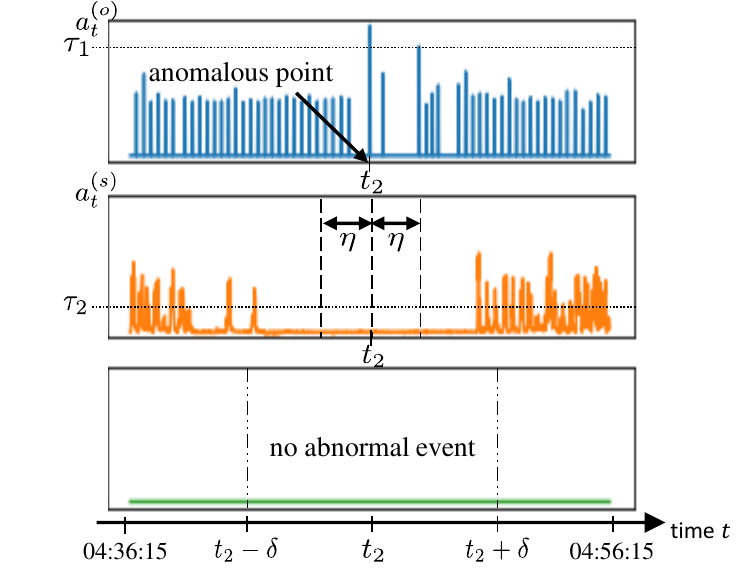}\label{fig:fig_2b}}
  \caption{An illustrative example of our T-DAD framework alongside two-stage signal-specific tasks.}
\label{fig:wrtL}
\end{figure}

Next, we turn to describing our two-stage signal-specific detection tasks. We optimally decide two thresholds $\tau_1$ and $\tau_2$ shown in (\ref{EQ:1}) and (\ref{EQ:3}), respectively, via exhaustive search in the sense of maximizing the performance (i.e., the $F_1$ score). To explain a rationale of our distinctive two-stage strategy, we give a motivating example as follows. Fig. \ref{fig:wrtL} illustrates the visualization of anomaly scores $a_t^{(o)}$ and $a_t^{(s)}$ (top and middle, respectively, of Figs. \ref{fig:fig_2a} and \ref{fig:fig_2b}) as well as actual abnormal events (bottom of Figs. \ref{fig:fig_2a} and \ref{fig:fig_2b}) within two different 20-minute fragments (05:30:22--05:50:22 on December 3, 2019 and 04:36:15--04:56:15 on December 7, 2019) in our time series dataset. From Fig. \ref{fig:fig_2a}, we make the following observations: 1) there is an anomalous point $t_1$ whose anomaly score $a_{t_1}^{(o)}$ is higher than the threshold $\tau_1$; 2) the maximum of anomaly scores $a_{t}^{(s)}, \forall t\in[t_1-\eta,t_1+\eta]$ is higher than another threshold $\tau_2$, which implies that the anomalous point $t_1$ will not be removed from the candidate; and 3) since the difference between the timestamp at which an actual abnormal event occurred and the \color{black}candidate point \color{black}$t_1$ is within the tolerance $\delta$, it is possible to successfully detect an actual abnormal event range-wisely. On the other hand, from Fig. \ref{fig:fig_2b}, one can see that 1) there is an anomalous point $t_2$ such that $a_{t_2} >\tau_1$; 2) the anomalous point $t_2$ is filtered out due to the fact that $\max_{t\in [t_2-\eta,t_2+\eta]} a_t^{(s)}<\tau_2$; and 3) no abnormal event is finally discovered. We summarize the overall procedure of our \color{black}T-DAD \color{black}framework in Algorithm 1.

 \begin{algorithm}[t]
 \caption{:~T-DAD}
 \begin{algorithmic}[1]
 \renewcommand{\algorithmicrequire}{\textbf{Input:}}
 \renewcommand{\algorithmicensure}{\textbf{Output:}}
 \REQUIRE ${\bf x}_t^{(o)}, {\bf x}_t^{(s)}, \tau_1, \tau_2, \eta$
 \ENSURE  $\hat{\mathcal{T}}_{\text{anomaly}}$
 \STATE {{\bf Initialization:} $\mathcal{T}_{o,\text{anomaly}}\leftarrow\emptyset, \mathcal{T}_s\leftarrow\emptyset$}
 \STATE {Train DAE with ${\bf x}_t^{(o)}$ on the training set}
 \STATE {Train LSTM-DAE with ${\bf x}_t^{(s)}$ on the training set}
 \FOR {$t \in \mathcal{T}_o$}
 \STATE {$a_t^{(o)} \leftarrow \|{\bf x}_t^{(o)}-\hat{\bf x}_{t}^{(o)}\|_2$}
 \IF {$a_t^{(o)} > \tau_1$}
 \STATE {$\mathcal{T}_{o,\text{anomaly}} \leftarrow \mathcal{T}_{o,\text{anomaly}} \cup \{t\}$}
 \ENDIF
 \ENDFOR
 \FOR {$t_i \in \mathcal{T}_{o,\text{anomaly}}$}
 \FOR {$t\in [t_i-\eta,t_i+\eta]$}
 \STATE {$a_t^{(s)} \leftarrow \max_{j \in J_t} \| {\bf x}_t^{(s)} - \hat{\bf x}_{t;j}^{(s)} \|_2$}
 \ENDFOR
 \IF {$\max_{t\in [t_i-\eta,t_i+\eta]} a_t^{(s)}<\tau_2$}
 \STATE {$\mathcal{T}_s \leftarrow \mathcal{T}_s \cup \{t\}$}
 \ENDIF
 \ENDFOR
 \STATE {$\hat{\mathcal{T}}_{\text{anomaly}} \leftarrow \mathcal{T}_{o,\text{anomlay}}\setminus \mathcal{T}_s$}
 \RETURN {$\hat{\mathcal{T}}_{\text{anomaly}}$}
 \end{algorithmic}
 \end{algorithm}

From the above empirical findings, we may conclude that 1) the DAE model based on operation cycle signals, aimed at detecting \color{black}highly likely \color{black}anomalous points, cannot solely solve the anomaly detection problem and 2) the LSTM-DAE model based on sensor signals plays a supplementary but crucial role in eliminating \color{black}very unlikely \color{black}anomalous points, thus resulting in much higher accuracy. 

\section{Experimental Evaluation}
\label{Sec:5}

In this section, we first describe the set of labels in our manufacturing dataset and data preprocessing techniques. Then, we introduce several performance metrics and five benchmark methods for anomaly detection. We also present experimental settings. As our main contributions, we comprehensively perform experimental \color{black}evaluations \color{black}to validate the superiority of our \color{black}T-DAD \color{black}framework over single-stage benchmark methods.

\subsection{Label Description}

True breakdown events caused by the assembly line are unavailable in our collected dataset; thus, in our experiments, the set of \color{black}alarm \color{black}points is used as a \color{black}ground truth \color{black}instead. In fact, many industrial plants include a key component for not only efficient and safe operation but also control of plants, the so-called alarm system~\cite{wang2015overview,goel2017industrial}, which alerts an operator about abnormal conditions or malfunctions of the process that might be \color{black}anomaly candidates\color{black}. Timely intervention by the operator alongside alarm logs in industrial datasets can be very useful in detecting possible malicious anomalies (i.e., abnormal events to possibly catastrophic levels).

In our dataset, an alarm point indicates the timestamp at which an alarm event arises and is recorded as a binary format. That is, if the value at a timestamp in a sequential time series of binary signals is 1, then it signifies that an alarm has occurred.

\subsection{Data Preprocessing}

We start by addressing that our dataset is collected from a total of 39 assembly processes. Among them, we focus primarily on the \nth{2} assembly process dataset for the analysis since it contains not only a number of features in each of heterogeneous signals but also a non-negligible fraction of alarm events used as a ground truth. There are 32 alarm points in the second assembly process dataset, which however contribute only to 0.06\% of all the timestamps. Additionally, the effectiveness of the proposed \color{black}T-DAD \color{black}framework will be empirically validated using other assembly process datasets.

Next, we apply the following three data preprocessing techniques into our selected assembly process dataset for successful analysis: data imputation, feature extraction, and data normalization. Specifically, we perform data imputation in such a way that the missing values in each signal are imputed with those observed at the previous timestamp. After the imputation of missing data, \color{black}we exclude some features such that their values remain unchanged over time. \color{black}This is because static features over time would be useless for data analysis. Finally, min-max normalization is adopted for each signal so that all the values in each signal are transformed into $[0,1]$, which enables us to alleviate the impact of different scales among the signals.

\subsection{Experimental Setup}

In this subsection, We first describe the settings of the neural network models used in our proposed \color{black}T-DAD \color{black}framework. Unless otherwise stated, as default settings, we run experiments using the DAE and LSTM-DAE models in Stages I and II, respectively, of \color{black}T-DAD \color{black}since this combination among the unsupervised learning models under consideration in Section V-E leads to the best performance (which will be specified in Section V-F).

In our experiments, both DAE models are trained using the Adam optimizer~\cite{kingma2014adam} with an initial learning rate of $10^{-3}$, where the batch size is set to 32. Loss functions based on the mean square error and the mean absolute error are used for the DAE and LSTM-DAE models, respectively. As for the DAE model, the MLP encoder--decoder has 2 hidden layers with the size of 6 each; the standard tanh and ReLU activation functions are applied alternately in every layer between input and output layers; an $\ell_1$-norm regularizer is adopted in the first layer of the encoder. As for the LSTM-DAE model, the encoder--decoder has 2 hidden LSTM layers with 64 hidden units each; dropout~\cite{JMLR:v15:srivastava14a} is applied to each layer with probability 0.2; the window size $w$ and the step size $\gamma$ are set to 180 and 60, respectively.

Additionally, in our experiments, the tolerance $\delta$ is set to 600 (seconds) since such a tolerance level may be acceptable in the sense of triggering an alert without causing a long delay. The range setting parameter $\eta$ used in Stage \RNum{2} of \color{black}T-DAD \color{black}is set to 14 (seconds) due to the fact that one operation cycle tends to be mostly from 13 to 15 seconds. Note that the parameters $\delta$ and $\eta$ can be set to different values according to the operator's criteria as well as assembly process environments.

We split our time series dataset into training (records measured from 00:00:00 on December 2, 2019 to 00:00:00 on December 3, 2019) and test (remaining records) sets in chronological order. During training, the two DAE models are learned; and the signal-specific detection tasks with threshold selection are conducted on the test set.

\subsection{Performance Metrics}

We adopt a modification of the well-known \color{black}Precision \color{black}and \color{black}Recall \color{black}so that we evaluate the performance of \color{black}range-wise \color{black}alarm detection appropriately since operators generally do not care about point-wise metrics, similarly as in~\cite{xu2018unsupervised,tatbul2018precision}. More specifically, in our study, the \color{black}Precision \color{black}means the fraction of finally detected anomalous points that find actual alarms within the tolerance $\delta$; and the \color{black}Recall \color{black}means the fraction of alarm points that are actually found by finally detected anomalous points within $\delta$. Furthermore, instead of the original $F_1$ score, which is the harmonic mean of Precision and Recall, given a particular threshold, we adopt the \color{black}best $F_1$ score \color{black}in~\cite{xu2018unsupervised,audibert2020usad,omnianomaly} that indicates the best possible performance of a model on the test set given the globally optimal threshold.\footnote{We remark that, under our dataset, it is hardly possible to properly tune the threshold values during training/validation since very few alarm points (used as labels) are available therein.} In our \color{black}T-DAD \color{black}framework, two thresholds are chosen in the sense of maximizing the $F_1$ score via exhaustive search, which would still be valuable in understanding the fundamental limits of unsupervised alarm detection in our manufacturing dataset.

\subsection{Benchmark Approaches}
\color{black}
We describe five benchmark methods for anomaly detection. They are used not only as backbone models in our two-stage framework but also as state-of-the-art single-stage benchmark approaches for comparison.

\begin{itemize}

    \item DAE~\cite{borghesi2019anomaly,zhou2017anomaly}: This model is the same as the Stage \RNum{1} model of our T-DAD framework in Section IV-C. 
    
    \item LSTM-DAE~\cite{malhotra2016lstm}: This model is the same as the Stage \RNum{2} model of our T-DAD framework in Section IV-C.

    \item DAGMM~\cite{dagmm}: Deep autoencoding Gaussian mixture model (DAGMM) aims at detecting anomalies by combining the AE and the Gaussian mixture model in an end-to-end fashion.

    \item LSTM-VAE~\cite{lstmvae}: This model uses a connected variational AE (VAE) with LSTM layers.
    
    \item USAD~\cite{audibert2020usad}: Unsupervised anomaly detection (USAD) utilizes adversely trained AEs inspired by GANs.
\end{itemize}
\color{black}

\begin{figure}[t]
    \centering
    \includegraphics[width=79mm]{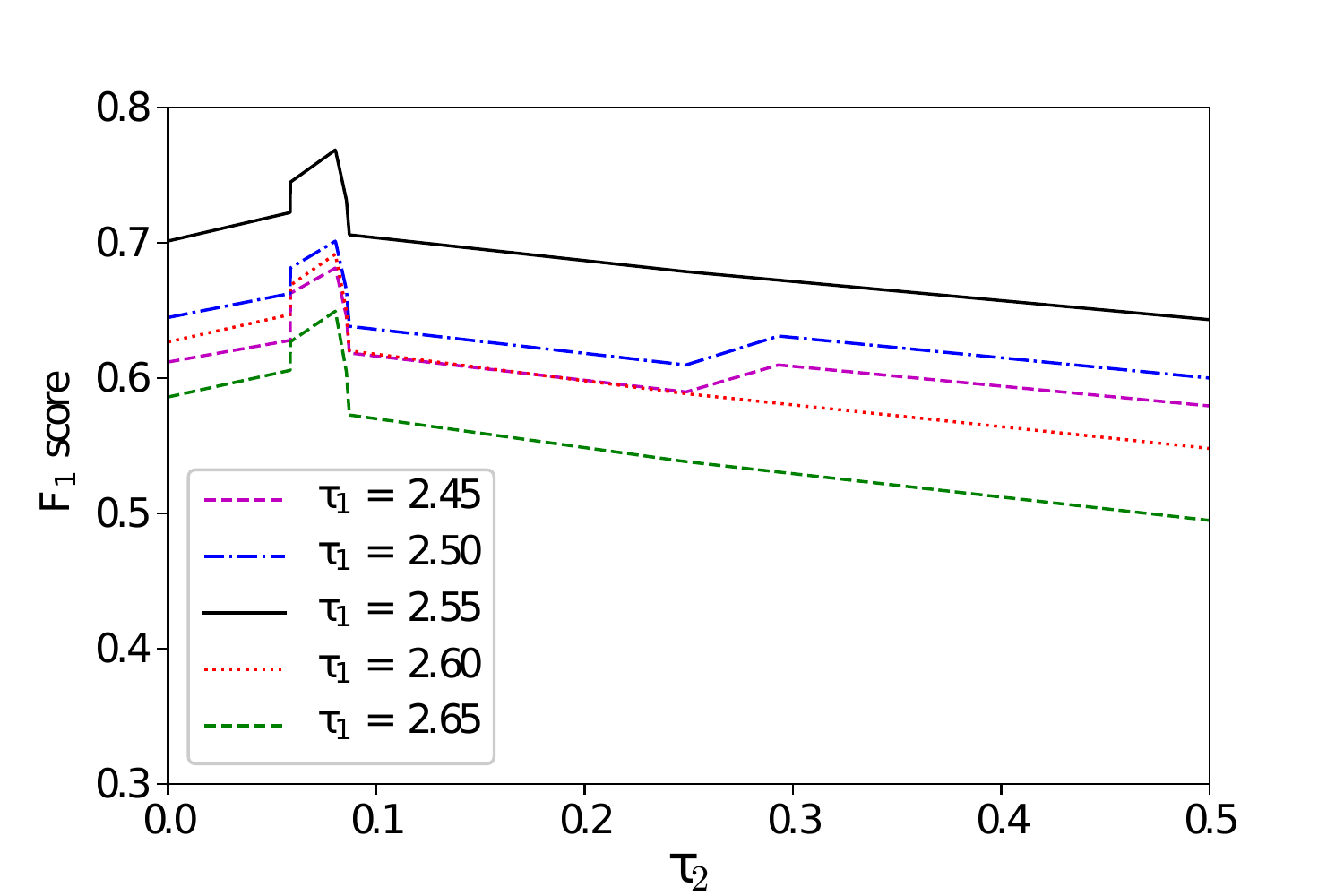}
    \caption{The $F_1$ score of T-DAD over two thresholds $\tau_1$ and $\tau_2$.}
    \label{fig:RQ1}
\end{figure}
 
\begin{table*}[ht]
\centering
\caption{Performance comparison between the T-DAD framework and single-stage benchmark methods.}
\label{tab:resultmain}
{\small
\def\arraystretch{1.3}
\begin{adjustbox}{max width=\textwidth}
% \resizebox{\textwidth}{!}{%
\begin{tabular}{|c|cccc|cccc|cccc|cccc|cccc|c|}
\hline
\multicolumn{1}{|l|}{}           & \multicolumn{20}{c|}{Single-stage methods}                                                                                              & Two-stage method \\ \hline
\multicolumn{1}{|c|}{Methods}    & \multicolumn{4}{c|}{DAE}  & \multicolumn{4}{c|}{DAGMM} & \multicolumn{4}{c|}{USAD} & \multicolumn{4}{c|}{LSTM-DAE} & \multicolumn{4}{c|}{LSTM-VAE} & {\bf {T-DAD}}                  \\ \hline
\multicolumn{1}{|c|}{Cases} & C1    & C2    & C3    & C4    & C1    & C2    & C3    & C4    & C1    & C2    & C3    & C4    & C1    & C2    & C3    & C4     & C1    & C2    & C3    & C4     & -              \\ \hline
{\em P}                        & 0.60  & 0.04 & 0.04 & 0.53 & 0.67  & 0.08 & 0.06 & 0.23 & 0.58 & 0.01 & 0.01 & 0.42 & 0.05  & 0.02  & 0.02  & 0.05  & 0.07  & 0.03  & 0.03  & 0.04  & 0.71                   \\ 
{\em R}                           & 0.84 & 0.69 & 0.44 & 0.66 & 0.56  & 0.34 & 0.97 & 0.41 & 0.78 & 0.06 & 0.06 & 0.72 & 0.25  & 0.59  & 0.5   & 0.25  & 0.18  & 0.56  & 0.63  & 0.13  &  0.84                   \\  
$F_1$                         & 0.70 & 0.08 & 0.07 & 0.59 & 0.61  & 0.13 & 0.10  & 0.29 & 0.67 & 0.01 & 0.02 & 0.53 & 0.09  & 0.04  & 0.03  & 0.09  & 0.10   & 0.06  & 0.06  & 0.06  & {\bf 0.77}                   \\ \hline
\end{tabular}%
\end{adjustbox}
% }
}
\end{table*}

\begin{table}[t]
\centering
\caption{Performance comparison among the model combinations $(\cdot, \cdot)$ in (Stage I, Stage II) of T-DAD}
\label{tab:resultagnostic}
{\small
\def\arraystretch{1.3}
\begin{adjustbox}{max width=\linewidth}
% \resizebox{\textwidth}{!}{%
\begin{tabular}{|c|ccc|ccc|ccc|}
\hline
\multirow{2}{*}{\backslashbox{Stage \RNum{2}}{Stage \RNum{1}}} & \multicolumn{3}{c|}{DAE}                      & \multicolumn{3}{c|}{DAGMM}                    & \multicolumn{3}{c|}{USAD}                     \\ \cline{2-10} 
                   & P             & R             & $F_1$            & P             & R             & $F_1$            & P             & R             & $F_1$            \\ \hline
% Single-stage          & 0.6           & 0.84          & 0.7           & 0.67          & 0.56          & 0.61          & 0.58          & 0.78          & 0.67          \\ 
DAE                & 0.60           & 0.84          & 0.7           & 0.67          & 0.56          & 0.61          & 0.58          & 0.78          & 0.67          \\ \
DAGMM              & 0.63          & 0.84          & 0.72          & 0.75          & 0.56          & 0.64          & 0.61          & 0.78          & 0.69          \\ 
USAD               & 0.63          & 0.84          & 0.72          & 0.75          & 0.56          & 0.64          & 0.63          & 0.78          & 0.69          \\ 
LSTM-DAE           & 0.71 & 0.84 & \bf{0.77} & 0.86 & 0.56 & \bf{0.68} & 0.65 & 0.78 & \bf{0.71} \\ 
LSTM-VAE           & 0.71 & 0.84 & \bf{0.77} & 0.86 & 0.56 & \bf{0.68} & 0.65 & 0.78 & \bf{0.71} \\ \hline
\end{tabular}%
\end{adjustbox}
% }
}
\end{table}

\subsection{Experimental Results}

\color{black}
In this subsection, our empirical study is designed to answer
the following four key research questions.
\color{black}

\begin{itemize}
    \item RQ1. How sensitive is the \color{black}T-DAD \color{black}framework to the selection of two thresholds $\tau_1$ and $\tau_2$?
    \color{black}
    \item RQ2. How much does the performance of T-DAD improve the detection accuracy over single-stage benchmark approaches?
    \item RQ3. Which combination of backbone models in T-DAD yields the best performance?
    \color{black}
    \item RQ4. How robust is the \color{black}T-DAD \color{black}framework to other assembly process datasets with \color{black}fewer \color{black}features?
\end{itemize}

To answer these questions, we carry out four comprehensive experiments as follows.

\subsubsection{Sensitivity Analysis with Respect to Thresholds (RQ1)}

To analyze the parameter sensitivity, we show the performance of the proposed \color{black}T-DAD \color{black}framework in Section~\ref{Sec:4} with respect to the $F_1$ score according to the values of thresholds $\tau_1$ and $\tau_2$. From Fig.~\ref{fig:RQ1}, it is observed that the performance tends to be more sensitive to the values of $\tau_1$ since a small variation in $\tau_1$ leads to a non-negligible difference in the $F_1$ score. One can also see that the best $F_1$ score is found at $(\tau_1,\tau_2)=(2.55,0.08)$, which will be used in the subsequent experiments unless otherwise specified.

\begin{table*}

\centering
\caption{Performance comparison on other assembly process datasets}
\label{table:table6}
\subtable[The \nth{5} assembly process dataset]{
{\scriptsize
\def\arraystretch{1.3}
\begin{adjustbox}{max width=\textwidth}
% \resizebox{\textwidth}{!}{%
\begin{tabular}{|c|cccc|cccc|cccc|c|}
\hline
\multicolumn{1}{|l|}{}           & \multicolumn{12}{c|}{Single-stage methods}                                                                                              & Two-stage method \\ \hline
\multicolumn{1}{|c|}{Methods}    & \multicolumn{4}{c|}{DAE} & \multicolumn{4}{c|}{DAGMM} & \multicolumn{4}{c|}{USAD} & {\bf T-DAD}                  \\ \hline
\multicolumn{1}{|c|}{Cases} & C1    & C2    & C3    & C4    & C1    & C2    & C3    & C4      & C1    & C2    & C3    & C4 & -              \\ \hline
{\em P} & 0.46  & 0.04 & 0.04 & 0.47 & 0.41 & 0.05 & 0.04 & 0.57 & 0.47 & 0.29 & 0.01 & 0.47 & 0.67                   \\ 
{\em R} & 0.71 & 0.35 & 0.35 & 0.41 & 0.71  & 0.65 & 0.53 & 0.38 & 0.71 & 0.15 & 0.24 & 0.71 & 0.62                   \\  
$F_1$ & 0.56 & 0.06 & 0.06 & 0.44 & 0.52 & 0.08 & 0.08  & 0.46 & 0.57 & 0.19 & 0.02 & 0.56 & {\bf 0.64}            \\ \hline
\end{tabular}%
\end{adjustbox}
% }
}
}
\subtable[The \nth{12} assembly process dataset]{
{\scriptsize
\label{table12th}
\def\arraystretch{1.3}
\begin{adjustbox}{max width=\linewidth}
% \resizebox{\textwidth}{!}{%
\begin{tabular}{|c|cccc|cccc|cccc|c|}
\hline
\multicolumn{1}{|l|}{}           & \multicolumn{12}{c|}{Single-stage methods}                                                                                              & Two-stage method \\ \hline
\multicolumn{1}{|c|}{Methods}    & \multicolumn{4}{c|}{DAE} & \multicolumn{4}{c|}{DAGMM} & \multicolumn{4}{c|}{USAD} & {\bf T-DAD}                  \\ \hline
\multicolumn{1}{|c|}{Cases} & C1    & C2    & C3    & C4  & C1    & C2    & C3    & C4   & C1    & C2    & C3    & C4     & -              \\ \hline
{\em P} & 0.14  & 0.06 & 0.10 & 0.13 & 0.10 & 0.09 & 0.08 & 0.08 & 0.13 & 0.15 & 0.15 & 0.12 & 0.43                   \\ 
{\em R} & 0.39 & 0.81 & 0.89 & 0.61 & 0.69  & 0.81 & 0.89 & 0.42 & 0.39 & 0.11 & 0.11 & 0.56 & 0.25                   \\  
$F_1$ & 0.20 & 0.11 & 0.18 & 0.21 & 0.11 & 0.17 & 0.16 & 0.15  & 0.19 & 0.13 & 0.13 & 0.19 & {\bf 0.32}               \\ \hline
\end{tabular}%
\end{adjustbox}
% }
}
}
\end{table*}

\subsubsection{Comparison with Single-Stage Benchmark Approaches (RQ2)}
\color{black}
Due to the fact that, to the best of our knowledge, anomaly detection exploiting all the heterogeneous signals in our real-world manufacturing dataset has never been studied yet, there is no method that works directly under our setting. For this reason, we demonstrate the superiority of our proposed two-stage framework over single-stage benchmark methods, where the DAE and LSTM-DAE models are used in Stages \RNum{1} and \RNum{2}, respectively, of our T-DAD framework. We take into account the following four cases as feasible single-stage approaches.

\begin{itemize}
    \item Case 1 (C1): Each method only uses operation cycle signals ${\bf x}_t^{(o)}, \forall t \in \mathcal{T}_o$.
    \item Case 2 (C2): Each method only uses sensor signals ${\bf x}_t^{(s)}$ for all timestamps.
    \item Case 3 (C3): We first impute missing values with zeros when operation cycle signals are not recorded. Then, each method uses the vector concatenation of both signals as input for all timestamps.
    \item Case 4 (C4): Each method uses the vector concatenation of operation cycle signals and sensor signals as input only when both signals are recorded, i.e., $[{\bf x}_t^{(o)}; {\bf x}_t^{(s)}] \in \mathbb{R}^{1\times (d_o+d_s)}, \forall t\in \mathcal{T}_o$.
\end{itemize}

We present the performance comparison between the T-DAD framework and five single-stage benchmark methods (including DAE, DAGMM, USAD, LSTM-DAE, and LSTM-VAE) for the above four cases in Table~\ref{tab:resultmain} with respect to the Precision, Recall, and best $F_1$ score (P, R, and $F_1$, respectively, for short in the table). We would like to provide the following insightful observations:

\begin{itemize}
    \item T-DAD is much superior to all single-stage benchmark approaches in terms of the best $F_1$ score;
    \item Only using operation cycle signals (i.e., C1) among four cases exhibits the best performance regardless of performance metrics, which reveals that operation cycle signals (rather than sensor signals) play a significant role in offering satisfactory performance;
    % \color{black}
    \item While the Recall of T-DAD is identical to that of C1 with DAE, the Precision of T-DAD is remarkably enhanced compared to the second best performer by virtue of judicious utilization of sensor signals in Stage II of T-DAD. Namely, sensor signals serve as a secondary tool in further improving the best $F_1$ score. This implies that our strategy in Stage II indeed leads to the removal of very unlikely anomalous points (i.e., reduction in false positives).
    % \color{black}
    \item Performance comparison between C1 and (C3, C4) indicates that the na\"ive vector concatenation of two signals even degrades the detection accuracy;
    \item From the comparison among five benchmark models in C1, it is seen that the best performer in terms of the best $F_1$ score is DAE while the second and third best performers are USAD and DAGMM, respectively;
    \item The performance of the LSTM-DAE and LSTM-VAE models is quite poor, which implies that simply using models with LSTM layers originally designed for learning temporal dependence rather deteriorates the detection accuracy.
\end{itemize}

\subsubsection{Model-Agnostic Property (RQ3)}

To show the model-agnostic property of our T-DAD framework, we diversify underlying models in Stages \RNum{1} and \RNum{2} (i.e., the Stage \RNum{1} and \RNum{2} models) by leveraging five benchmark methods shown in Section V-E. In Stage \RNum{1} where operation cycle signals are used, we choose DAE, DAGMM, and USAD for model candidates since using the other two (i.e., LSTM-DAE and LSTM-VAE) do not perform appropriately as stated in Section V-F2). On the other hand, in Stage II where sensor signals are used, we choose all five methods as model candidates. 

The performance comparison among the model combinations $(\cdot, \cdot)$ in (Stage I, Stage II) of T-DAD is presented in Table~\ref{tab:resultagnostic} with respect to the Precision, Recall, and best $F_1$ score (P, R, and $F_1$ for short in the table). Our findings are as follows. First, experimental results clearly exhibit the model-agnostic property of our T-DAD framework while showing that the adoption of two stages indeed enhances the performance over its counterpart (i.e., single-stage benchmark methods). From Tables~\ref{tab:resultmain} and \ref{tab:resultagnostic}, note that the performance of the single-stage approach showing the highest best $F_1$ score among four cases (i.e., C1) is identical to that of using DAE in Stage II of T-DAD regardless of models in Stage I. Second, it is examined that RNN-based models such as LSTM-DAE and LSTM-VAE are more appropriate in Stage II than other benchmark methods due to the fact that sensor signals tend to show strong temporal dependence. From Table~\ref{tab:resultagnostic}, we may conclude that either (DAE, LSTM-DAE) or (DAE, LSTM-VAE) is the best combination of models in our T-DAD framework.

\color{black}

\subsubsection{Robustness to Fringe Conditions (RQ4)}

To validate the robustness of our \color{black}T-DAD \color{black}framework, we perform anomaly detection on \color{black}two other \color{black}assembly process datasets whose characteristics are appropriate for our experimental study in terms of not only a balanced number of features in each of heterogeneous signals but also a non-negligible fraction of alarms. Specifically, we note that 32 out of 39 assembly processes contain only a very few operation cycle or sensor signals, which means that our framework exploiting heterogeneous time series signals may be unnecessary (i.e., adoption of an existing single-stage detection model rather than multi-stage methods would be satisfactory in such cases). Among the remaining 7 assembly process datasets, assembly process datasets that have less than 30 alarms are excluded from our experiments in order to avoid unreliable verification caused by the very scarce labels. In consequence, we choose the following two assembly process datasets:
\begin{itemize}
    \item The \nth{5} assembly process dataset having 22 features in operation cycle signals and 7 features in sensor signals after feature extraction;
    \item The \nth{12} assembly process dataset having 11 features in operation cycle signals and 6 features in sensor signals after feature extraction.
\end{itemize}

\color{black}
To demonstrate the validity of our T-DAD framework on other datasets, we present the performance comparison between T-DAD and three single-stage benchmark approaches (including DAE, DAGMM, and USAD) in Table~\ref{table:table6}, where the DAE and LSTM-DAE models are used in Stages I and II, respectively, of our T-DAD framework. We do not adopt the other two benchmark methods (i.e., LSTM-DAE and LSTM-VAE) since they do not perform appropriately similarly as in Section VF2). It is clear to see that the results in the table follow similar trends to those in Table~\ref{tab:resultmain}. On both assembly process datasets, T-DAD shows superior performance over single-stage benchmark approaches, which demonstrates that our T-TAD framework is robust to other assembly process datasets.
\color{black}

\section{Concluding Remarks}
\label{Sec:6}

In this paper, we introduced a novel two-stage framework, termed T-DAD, for automatically detecting anomalies using a real-world manufacturing dataset, collected from an assembly line of automotive engine valves, containing many challenging problems. \color{black}More precisely, we developed our T-DAD framework in such a way that signal-specific detection tasks are carried out in a data-driven manner after unsupervisedly training a model in stage \RNum{1} with operation cycle signals and another model in stage \RNum{2} with sensor signals. Through comprehensive experiments, we demonstrated that the proposed framework not only substantially outperforms various single-stage benchmark methods \color{black}but also is robust to fringe situations. Moreover, we found that the gains over single-stage benchmark methods come from our two-stage detection strategy depending on types of signals by judiciously exploiting sensor signals as a secondary tool in further improving the performance. Potential avenues of future research include how to solve a general setting where we undergo more heterogeneity along with more than three types of signals in highly sophisticated real-world manufacturing. Early detection of anomalies for preventative maintenance also remains for future work.

% \section*{Acknowledgment}
% The material in this paper has been presented in part at the IEEE International Conference on Communications, Paris, France, May 2017 \cite{lin2017achieving} and the IEEE International Symposium on Information Theory, Aachen, Germany, June 2017 \cite{lin2017multi}.

% \bibliographystyle{ieeeaccess}
% \bibliographystyle{unsrt}

\bibliographystyle{IEEEtran}
\bibliography{IEEEabrv,References}
\end{document}